\documentclass[runningheads]{llncs}
\usepackage[T1]{fontenc}
\usepackage{graphicx}
\usepackage{cite}
\usepackage{amsmath,amssymb,amsfonts}
\usepackage{algorithmic}
\usepackage{graphicx}
\usepackage{textcomp}
\usepackage{amsmath}
\usepackage{amssymb}
\usepackage{hyperref}
\usepackage{booktabs}
\usepackage{svg}
\usepackage{adjustbox}
\usepackage{graphicx}
\usepackage{csvsimple}
\usepackage{siunitx}
\usepackage{caption}
\usepackage{subcaption}
\usepackage{pgffor}
\usepackage{upgreek}
\usepackage{mathtools}
\usepackage{float}
\usepackage{listings}
\usepackage{xcolor}
\usepackage[normalem]{ulem}
\usepackage{wrapfig}
\usepackage{orcidlink}
\usepackage{newtxtext,newtxmath}
\usepackage{bbding}
\usepackage{hyperref}
\hypersetup{colorlinks,urlcolor=blue}

\newcommand\blfootnote[1]{%
  \begingroup
  \renewcommand\thefootnote{}\footnote{#1}%
  \addtocounter{footnote}{-1}%
  \endgroup
}

\makeatletter
\DeclareUrlCommand\ULurl@@{%
  \def\UrlLeft{\uline\bgroup}%
  \def\UrlRight{\egroup}}
\def\ULurl@#1{\hyper@linkurl{\ULurl@@{#1}}{#1}}
\DeclareRobustCommand*\ULurl{\hyper@normalise\ULurl@}
\makeatother

\begin{document}
\title{Visualization and Analysis of the Loss Landscape in Graph Neural Networks}
%
%
\author{Samir Moustafa\inst{1,2}$^($\Envelope$^)$\orcidlink{0000-0002-0674-9667} \and
Lorenz Kummer\inst{1,2}\orcidlink{0000-0001-6538-9107} \and
Simon Fetzel\inst{1} \and
Nils M. Kriege\inst{1,3}\orcidlink{0000-0003-2645-947X} \and
Wilfried N. Gansterer\inst{1}\orcidlink{0000-0001-5170-1251}}
\authorrunning{S. Moustafa et al.}
%
\institute{Faculty of Computer Science, University of Vienna, Vienna, Austria \and
UniVie Doctoral School Computer Science, University of Vienna, Vienna, Austria \and
Research Network Data Science, University of Vienna, Vienna, Austria \\
\email{\{samir.moustafa, lorenz.kummer, simonf93, nils.kriege, wilfried.gansterer\}@univie.ac.at}\\
}
\maketitle              
\begin{abstract}

Graph Neural Networks (GNNs) are powerful models for graph-structured data, with broad applications. However, the interplay between GNN parameter optimization, expressivity, and generalization remains poorly understood. We address this by introducing an efficient learnable dimensionality reduction method for visualizing GNN loss landscapes, and by analyzing the effects of over-smoothing, jumping knowledge, quantization, sparsification, and preconditioner on GNN optimization.
Our learnable projection method surpasses the state-of-the-art PCA-based approach, enabling accurate reconstruction of high-dimensional parameters with lower memory usage.
We further show that architecture, sparsification, and optimizer's preconditioning significantly impact the GNN optimization landscape and their training process and final prediction performance.
These insights contribute to developing more efficient designs of GNN architectures and training strategies.

\blfootnote{This is a preprint of the work accepted to the International Conference on Artificial Neural Networks Workshops, LNCS 16072, pp. 1–13. \url{doi.org/10.1007/978-3-032-04552-2\_9}}.
\end{abstract}

\section{Introduction}
Graph Neural Networks (GNNs) are tailored for graph-structured data and excel in tasks like network analysis, molecular property prediction, and recommendation systems~\cite{LingfeiGNN}. Compared to Deep Neural Networks (DNNs), they have fewer parameters but higher computational costs due to large input graphs~\cite{Moustafa2025EfficientMP, Zhu2023rmAA}. The relationship between GNNs optimization dynamics and graph structure, architectural design, or numerical precision is poorly understood~\cite{LingfeiGNN}.

\begin{figure}
    \footnotesize
    \begin{subfigure}{0.24\textwidth}
    \includegraphics[width=\textwidth]{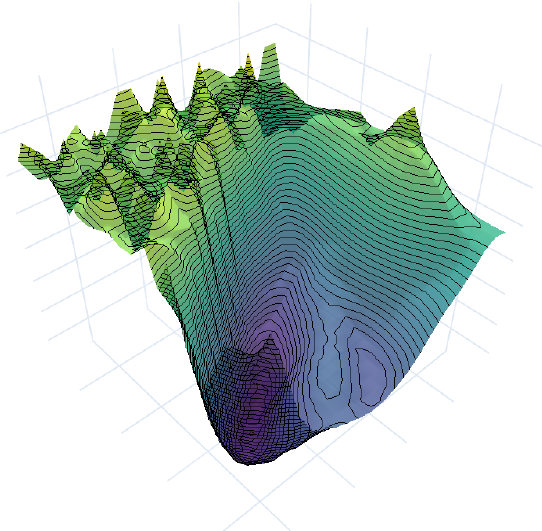}
    \caption{\footnotesize ResNet56 without skip connections}
    \label{fig:resnet_no_skip}
    \end{subfigure}
    \hfill
    \begin{subfigure}{0.24\textwidth}
    \includegraphics[width=\textwidth]{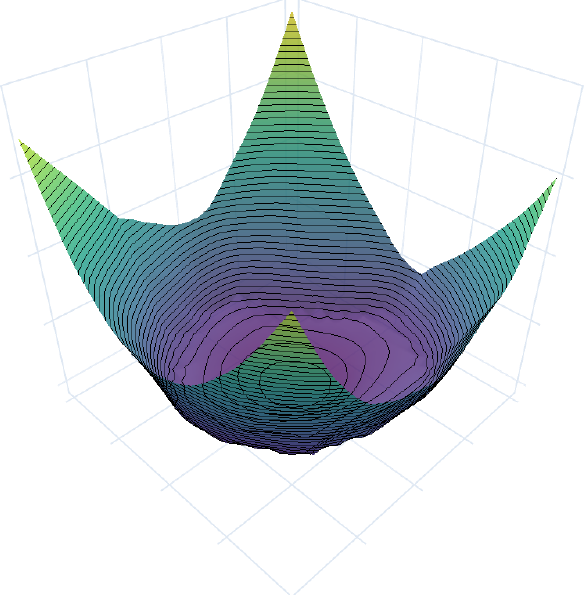}
    \caption{\footnotesize ResNet56 with skip connections}
    \label{fig:resnet_with_skip}
    \end{subfigure}
    \begin{subfigure}{0.24\textwidth}
    \includegraphics[width=\textwidth]{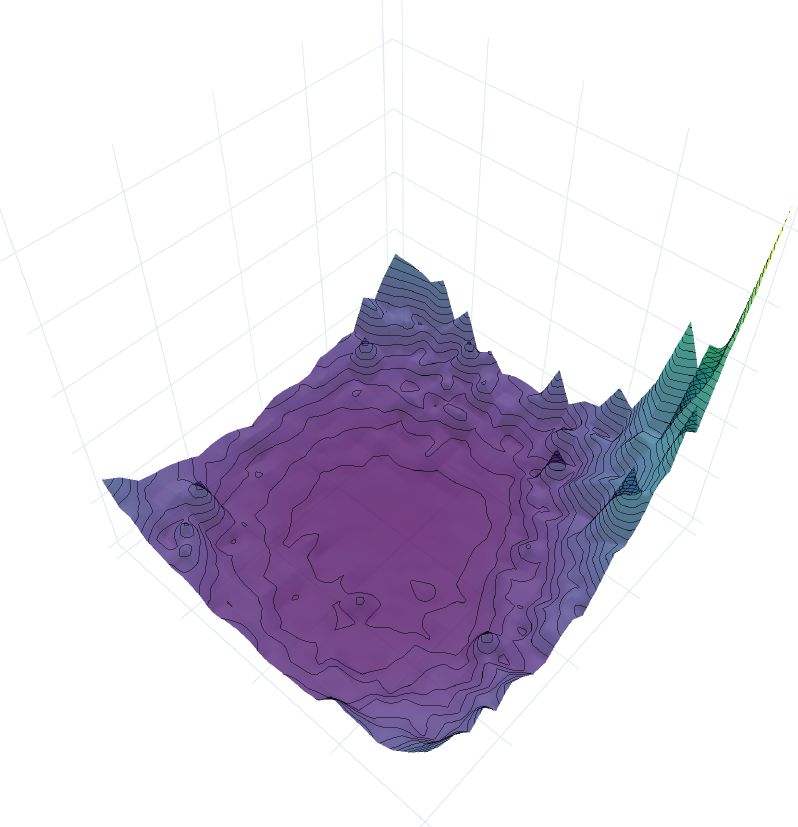}
    \caption{\footnotesize GCN without jumping knowledge}
    \label{fig:gcn_no_jk}
    \end{subfigure}
    \hfill
    \begin{subfigure}{0.24\textwidth}
    \includegraphics[width=\textwidth]{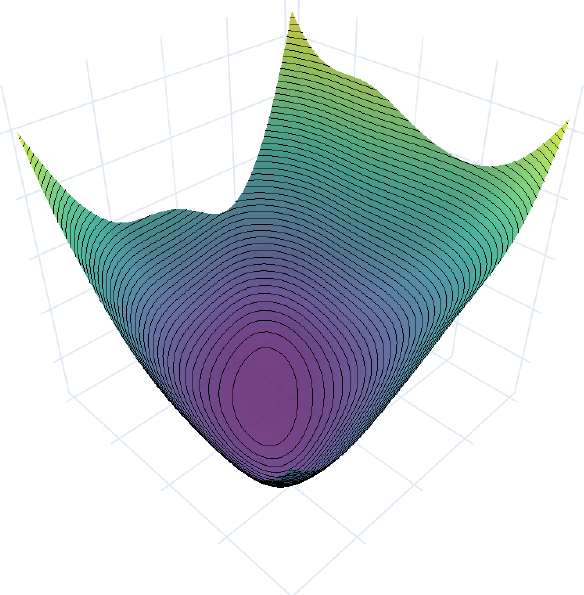}
    \caption{\footnotesize GCN with jumping knowledge}
    \label{fig:gcn_with_jk}
    \end{subfigure}
    \caption{
    3D loss landscapes of ResNet56 on CIFAR-10~\cite{KrizhevskyCIFAR10} and GCN on Cora~\cite{Giles1998CiteSeerAA}, with and without architectural modifications.}
    \label{fig:3d_skip_connections_vs_jumping_knowledge}
\end{figure}

GNN research has focused chiefly on theoretical expressivity--the ability to distinguish non-isomorphic graphs or structurally distinct nodes~\cite{WLsurvey2021}. While deeper networks are more expressive, over-smoothing~\cite{LingfeiGNN} limits practical depth by making node embeddings indistinguishable. Although theoretical expressivity results guarantee the existence of suitable parameters, standard optimization techniques may fail to recover them, revealing a gap between theory and practice. The interplay among expressivity, generalization, and optimization remains insufficiently understood~\cite{WLsurvey2021}.

Analyzing loss landscapes and training trajectories can illuminate the gap between theoretical and practical GNN behavior, yet it remains underexplored. The loss landscape maps parameters to loss values, offering a geometric view of model behavior, and is typically highly non-convex and complex, particularly under varying data distributions~\cite{Li2017visualizing}. Understanding the loss landscape yields insights into optimization, generalization, and the stability of learned representations~\cite{Choromanska15}. Loss landscape analysis for GNNs remains unexplored, hindered by challenges inherent to graph-structured data, message passing, and architectural diversity~\cite{LingfeiGNN}. Recent advancements in GNNs training techniques, aimed at improving generalization~\cite{Xu2018}, reducing inference time~\cite{graphsaint-iclr20,Zhu2023rmAA}, or optimizing parameters~\cite{izadi2020optimization}, have further complicated the understanding of the GNN loss landscape by obscuring the relationship between model parameters and loss.

This paper addresses this gap and sheds light on the importance and methods of analyzing and visualizing the loss landscape in GNNs.

\paragraph{Contribution of the Paper:}
Our contributions are threefold:
\textbf{(1)} We propose a novel learnable projection technique for loss landscape visualization, mapping high-dimensional parameters into 2D and restoring them, while supporting batching for controlling memory usage.
\textbf{(2)} We provide a comprehensive analysis of GNN training techniques and their effects on the loss landscape, considering over-smoothing~\cite{LingfeiGNN}, jumping knowledge~\cite{Xu2018}, quantization~\cite{Moustafa2025EfficientMP}, sparsification~\cite{graphsaint-iclr20}, and optimizer preconditioning~\cite{izadi2020optimization}.

\paragraph{Preliminaries:}
\label{sec:preliminaries}
A \emph{graph} $G$ is a pair $(V,E)$, where $V$ is a finite set of nodes and $E$ is a finite set of edges. Let $A$ denote the adjacency matrix of $G$ with order $|V|$. Each node $v_i \in V$ has an associated feature vector $x_i \in \mathbb{R}^{f}$, stacked into a matrix $X \in \mathbb{R}^{|V| \times f}$. We consider a classification setting, assuming each node has an expected class, and in total there are $C$ classes, represented by $Y \in \mathbb{R}^{|V| \times C}$.

A GNN is parameterized by $\uptheta$, defining a function $\mathcal{F}_\uptheta: \left(A, X\right) \rightarrow Y$. GNNs employ message passing, where each node aggregates neighbor embeddings, implemented via multiplication of the adjacency matrix $A$ and feature matrix $X$. At the $k^\text{th}$ layer, $X^{(k)}$ is computed as $A X^{(k-1)} \uptheta^{(k)}$, where $X^{(k-1)}$ is the embedding from layer $k-1$~\cite{LingfeiGNN}.
Training of a GNN aims to adjust $\uptheta$ so that $\mathcal{F}_\uptheta(X)$, given $G$, approximates the target $Y$. During training, the loss function $\varepsilon$ measures the discrepancy between the expected and actual outputs; we use multi-class cross-entropy, though any differentiable loss is applicable. During training, gradients of $\varepsilon(\mathcal{F}_\uptheta(A, X), Y)$ with respect to $\uptheta$ are computed to update parameters and minimize the error.

A loss landscape visualization depicts how the loss varies with the model parameters.
Ideally, a well-fitted model yields a smooth, convex landscape, as shown in Figures~\ref{fig:resnet_with_skip} and~\ref{fig:gcn_with_jk}.

\paragraph{GNN Architectures:}
This work focuses on three fundamental GNN architectures selected for their pivotal role in the evolution and comprehension of GNNs.
Graph Convolution Network (GCN)~\cite{Kipf2016} applies convolution in the spectral domain to capture local graph structures invariant to graph isomorphisms.
Graph Attention Network (GAT)~\cite{velickovic2018graph} utilizes an attention mechanism that allows nodes to weigh their neighbors' influence dynamically, enhancing model flexibility and performance on graph-structured data.
Graph Isomorphism Network (GIN)~\cite{Xu2018a} is more powerful in distinguishing different graph structures than GCN and GAT by using a multilayer perceptron (MLP) and a learnable parameter for feature updating, achieving theoretical equivalence with the Weisfeiler-Lehman test. Table~\ref{tab:gnn_architecture} presents the mathematical formulations for GCN, GIN, and GAT, detailing aggregation and update functions based on hidden states $h_i$, output embeddings $x_i$, neighborhood structures $\mathcal{N}(i)$, and the $\alpha_{ij} = \frac{e_{i,j}}{\sum_{k\in\mathcal{N}(i)} e_{i,k}}$, such that $e_{i,j}=\mathscr{e}^{\text{LeakyReLU}\left(a^\top[\theta h_i \|\theta h_j]\right)}$, and $a$ is the attention learnable parameter.

\begin{table}[t]
    \setlength{\tabcolsep}{1em}
    \centering
    \caption{Standard aggregation and update functions in the $k^\text{th}$ layer of three GNN architectures: GCN, GAT, and GIN. These functions gather (aggregation) and transform (update) operations by collecting information from each node’s neighborhood.}\vspace{0.35em}
    \resizebox{\linewidth}{!}{
        \begin{tabular}{lll}
        \toprule[2pt]
        \textbf{\small Architecture} & \textbf{\small Aggregation Function} & \textbf{\small Update Function} \\
        \midrule[1pt]
        GCN~\cite{Kipf2016} & $h_i^{(k)}=\sum\limits_{j \in \mathcal{N}(i) \cup\{i\}} \frac{1}{\sqrt{d_i d_j}} x_j^{(k-1)}$ & $x_i^{(k)}=\operatorname{ReLU}\left(\uptheta^{(k)} h_i^{(k)}\right)$ \\ \\
        GAT~\cite{velickovic2018graph} & $h_i^{(k)}=\sum\limits_{j \in \mathcal{N}(i) \cup\{i\}} \alpha_{i, j}^{(k)} x_j^{(k-1)}$ & $x_i^{(k)}=\operatorname{ReLU}\left(\uptheta^{(k)} h_i^{(k)}\right)$ \\ \\
        GIN~\cite{Xu2018a} & $h_i^{(k)}=\left(1+\epsilon^{(k)}\right) x_i^{(k-1)}+\sum\limits_{j \in \mathcal{N}(i)} x_j^{(k-1)}$ & $x_i^{(k)}= \operatorname{MLP}^{(k)}\left(h_i^{(k)}\right)$ \\
        \bottomrule[2pt]
    \end{tabular}
    }
    \label{tab:gnn_architecture}
\end{table}

\paragraph{Problem Definition:}
The loss landscape $L(\uptheta)$ depends on the architecture, input data, target output, parameters, and error function. Visualizing $L(\uptheta)$ helps reveal how the error adapts to parameter changes, where $L(\uptheta)$ reduces to $\varepsilon(\mathcal{F}_{\uptheta}(A, X), Y)$ with other factors fixed. This landscape captures the model’s error across parameter configurations. Optimization seeks parameters minimizing $\varepsilon$, but is hindered by non-convexity and local minima. Due to the high dimensionality of $\uptheta$, a full mathematical analysis of $L(\uptheta)$ is typically intractable.

\section{Related Work}
\label{sec:related_work}
In this section, we delve into the definition of the loss landscape, the limitations of GNNs, explore various methods that have been adapted specifically for GNNs, and discuss the prevalent optimization challenges associated with them. 

\paragraph{Loss Landscape Visualization:}
Despite the non-convexity of $L(\uptheta)$, DNNs often train efficiently by converging to local minima. This raises questions~\cite{Goodfellow2014Qualitatively} about whether training avoids local minima or if saddle points dominate optimization dynamics. \cite{Goodfellow2014Qualitatively} study $L(\lambda , \uptheta_i + (1 - \lambda) , \uptheta^\ast)$ for $\lambda \in \mathbb{R}$, where $\uptheta_i$ and $\uptheta^\ast$ are parameters from epoch $i$ and the final training epoch, respectively. They observe elevated loss regions between solutions, consistent with~\cite{Im2016Empirical}. Visualization techniques include barycentric and bilinear interpolation, and projections along random or PCA directions~\cite{Lorch2016,Li2017visualizing}.
Moreover,~\cite{Im2016Empirical} analyzed how optimization trajectories change when the algorithm switches mid-training, typically after loss stagnation. Post-switch trajectories consistently diverge, suggesting optimizers follow distinct paths at saddle points. Loss landscapes remain similar across runs with the same optimizer but different initializations. Despite differing final parameters, loss and accuracy are often comparable—a result also observed in~\cite{Smith2017ExploringLF}, where varying the learning rate led to multiple equally performant optima~\cite{Hochreiter97FlatMinima}.

\paragraph{Over-smoothing and Jumping Knowledge:}
A key challenge in GNN training is over-smoothing, where node embeddings become increasingly similar with depth, diminishing node identity and discriminative power~\cite{LingfeiGNN}. Over-smoothing occurs as the row of $X^{(l)}$ converges to the same vector as $l \to \infty$~\cite{LingfeiGNN}. This can be quantified via similarity or distance measures between rows of $X^{(l)}$.

Jumping Knowledge (JK) mitigates over-smoothing by allowing node representations to aggregate multi-distance neighborhood information across layers. The final representation is given by $X^{(l)} = \phi(X^{(0)}, X^{(1)}, ..., X^{(l-1)})$, where $\phi$ combines outputs from all previous layers. JK parallels residual connections in CNNs~\cite{KaimingDeepResidual}, enhancing trainability and addressing vanishing gradients.

\paragraph{Quantization and Sparsification:}
Quantization lowers memory, computation, and power demands by converting weights or activations to lower-precision formats, enabling efficient deployment at the cost of approximation errors~\cite{gholami2022survey}. \cite{Zhu2023rmAA} proposes a GNN-specific method that learns per-node bit-widths based on aggregation values, capturing topological variance. This adds complexity, requiring group-specific learning rates.

Sparsification reduces GNN memory and computation by constructing sparse graphs that retain essential structure. GraphSAINT~\cite{graphsaint-iclr20} samples subgraphs to preserve batch connectivity, using normalization and diverse strategies to control bias and variance. Its effectiveness depends critically on sampling quality~\cite{LingfeiGNN}.

\paragraph{Optimization Challenges and Strategies:}
\label{sec:optimization}
DNN optimization has shifted focus from local minima, which may support generalization~\cite{Hochreiter97FlatMinima}, to saddle points as the main challenge~\cite{Im2016Empirical, Saxe2013ExactST}. Low-error solutions exhibit parameter space symmetry, with local minima often matching global optima in fully connected networks~\cite{Nguyen2017, Choromanska15}. Under conditions like identity mappings or over-parameterization, stochastic gradient descent (SGD) can reach global optima~\cite{Li2017, Soltanolkotabi2017}. In high dimensions, saddle points prevail, and methods like natural gradient descent (NGD) are effective for escaping them~\cite{Dauphin2014}.
\cite{izadi2020optimization} introduces an information-geometric optimization method for GNNs using NGD, with the Fisher Information Matrix approximated via KFAC to avoid second-order derivatives. This improves efficiency and outperforms ADAM and SGD.

Overall, the unique structural characteristics of GNNs, alongside their phenomena and methodologies, present unexplored dimensions in understanding how they shape the loss landscape and govern optimization trajectories.

\section{Methodology}
Consider a $k$-layer graph neural network with parameters $\uptheta = (\theta^0, \theta^1, \dots, \theta^k)$, where each $\theta^i$ is a vector or matrix. Let $\Theta \in \mathbb{R}^d$ denote the flattened concatenation of all elements in $\uptheta$, with $d$ as the parameter dimension. The optimized flattened concatenated parameters are denoted $\Theta^\ast \in \mathbb{R}^d$.

\paragraph{Dimensionality Reduction for Visualization:}
To visualize the high-dimensional parameter space $\mathbb{R}^d$, dimensionality reduction to 2D is applied using two directions $b^1, b^2 \in \mathbb{R}^d$. The mapping from 2D to $\mathbb{R}^d$ is defined by the function
\begin{align}
    \label{equ:mapping}
    M: \; \mathbb{R}^2 \rightarrow \mathbb{R}^d \text{ with }
     (x,y) \mapsto \Theta^\ast + x\cdot b^1 + y\cdot b^2,
\end{align}
where $x$ and $y$ are the visualization's axes.
The visualization is then created via 
\begin{equation}
    \label{equ:demapping}
    \Psi: (x,y) \mapsto L(M(x,y))
\end{equation}
The directions $b^1, b^2$ can be selected via various methods; one common approach is random initialization~\cite{Li2017visualizing}. Each direction vector $(b_0, b_1, \dots, b_d)$ is partitioned into segments $\hat{b}_l = (b_m, \dots, b_k)$, corresponding to specific network parameters (e.g., weight matrices or bias vectors). These segments are normalized using their magnitude and that of the corresponding optimal parameters $\hat{\Theta}^\ast_l = (\Theta^\ast_m, \dots, \Theta^\ast_k)$, ensuring each direction is scaled relative to its associated optimized values.
This scaling ensures that each direction vector segment matches the magnitude of its corresponding optimal parameter segment. The normalization is formalized as:
\begin{equation}
    \label{equ:normalization}
    \hat{b}_j' \coloneqq \frac{\hat{b}_l}{\lVert \hat{b}_l \rVert} \: \lVert \hat{\Theta}^\ast_{l} \rVert
\end{equation}
Normalized random directions allow a more balanced and representative visualization of the high-dimensional parameters.

\paragraph{Visualizing the Optimizer Trajectory:}
\label{sec:optimizer_trajectory}
To visualize optimizer trajectories, directions $b^1$ and $b^2$ are selected to align with the subspace spanned by parameters $\Theta_i$, $0\leq i \leq n$, across all $n$ epochs. Using $\Theta^\ast$ as the origin, differences $D_i = \Theta_i - \Theta^\ast$ form a matrix $D \in \mathbb{R}^{n \times d}$.
PCA on the covariance of $D$ yields $b^1$ and $b^2$~\cite{Lorch2016}. However, computing the covariance matrix and its eigenvectors is memory-intensive for large networks\cite{Li2017visualizing}, despite $n \ll d$ reducing the impact of $D$ itself.

\paragraph{Projection into Visualization Space:}
\label{sec:projection}
To visualize trajectories, a point $p \in \mathbb{R}^d$ must be projected into the 2D visualization space $(x, y)$ by solving:
\begin{equation}
    \label{equ:projection}
    p=\Theta^\ast + x\cdot b^1 + y\cdot b^2 + r
\end{equation}
where $r$ is the reconstruction error vector, which should be minimized.
Basis pairs can be compared via their reconstruction error $r$, whereby a smaller $r$ indicates a preferable basis. If the basis vectors $b^1,b^2$ are orthonormal~\cite{Li2017visualizing}, the dot product can be used to calculate the coordinates as
$x = \langle p - \Theta^\ast, b^1 \rangle$  and
$y = \langle p - \Theta^\ast, b^2 \rangle$.
In general, $b^1, b^2$ are not orthonormal and not orthogonal. Therefore, an underdetermined linear equation system needs to be solved.
\begin{equation}
    \begin{pmatrix}
    b^1 & b^2
    \end{pmatrix}
    \begin{pmatrix}
    x\\
    y
    \end{pmatrix}
    =
    p - \Theta^\ast
\end{equation}
The solution only approximates $p$; the true loss may differ from $\Psi(x, y)$ in equation~(\ref{equ:demapping}). Hence, 2D trajectory visualizations such as contour plots are preferred.

\paragraph{Learnable Projection:}
\label{sec:learnable_projection}
We propose a novel dimensionality reduction method, the \emph{learnable projection model}, defined by a matrix $P \in \mathbb{R}^{2 \times d}$, where $P = \left({b^1} \ {b^2}\right)^\top$. $P$ defines the projection unambiguously, encodes high-dimensional parameters into 2D, and decodes them back. Given input matrix $D$ from Section~\ref{sec:optimizer_trajectory}, each point is encoded as $z_i = D_i P^\top \in \mathbb{R}^2$, assuming orthonormal $b^1, b^2$ (see Section~\ref{sec:projection}). Decoding is performed via $z_i P$, equivalent to the mapping function $M$ in equation~(\ref{equ:mapping}).

The learnable projection model is trained to minimize the Euclidean distance between $D$ and $zP$ via Mean Squared Error (MSE), thereby minimizing the reconstruction error $r$ from equation~(\ref{equ:projection}) and yielding a 2D basis with low projection error.
For optimization, the problem is formulated as:
\begin{equation}
\min_{b^1,b^2\in \mathbb{R}^d} 
\;\mathrm{MSE}\bigl(D,\,D P^\top P\bigr), 
\quad D\in\mathbb{R}^{n\times d}, \quad
P = \left({b^1} \ {b^2}\right)^\top \in \mathbb{R}^{2\times d}.
\label{equ:learnable_optimization}
\end{equation}
Simplifying the objective yields an upper bound: minimizing ${\lVert I_d - P^\top P \rVert}_2^2$, where $I_d$ is the $d \times d$ identity matrix. This reformulation shifts the focus to minimizing the squared spectral norm of $I_d - P^\top P$.
This formulation relates to the orthogonal Procrustes problem, though it omits explicit dependence on $\Theta^\ast$ and $\Theta_i$ encoded in $D$. The $\ell_1$ reconstruction error $\lVert r \rVert_1$ enables comparison with PCA-based methods, as both aim to minimize the same error term from equation~(\ref{equ:projection}), 
which highlights the effectiveness of our learnable projection relative to PCA, despite both sharing the same objective.

Theoretically, both our approach and PCA aim to minimize the \textit{reconstruction error}, as defined in equation~\ref{equ:reconstruction_error} from~\cite{Hastie09}:
\begin{equation}
    \min_{V_q}\sum_{i=1}^N ||\left(x_i-\overline{x}-V_qV_q^\top(x_i-\overline{x})\right)||^2
    \label{equ:reconstruction_error}
\end{equation}
where $x_i$ are the data samples, $\overline{x}$ is the mean and $V_q$ is a matrix consisting of $q$ orthonormal columns. 
Unlike PCA, which computes $V_q$ via eigenvectors, our approach uses gradient-based optimization, avoiding covariance matrix computation and associated memory overhead.
Grouping rows of $D$ into batches enables the learnable projection to train on $D_i$ in a batched manner, reducing memory usage even further. 

\begin{table}[!t]
    \centering
    \caption{Characteristics of node classification datasets across different domains.}
    \resizebox{\linewidth}{!}{
    \begin{tabular}{crrrrr}
    \toprule[2pt]
    & \textbf{Cora}~\cite{Giles1998CiteSeerAA} & \textbf{Citeseer}~\cite{Giles1998CiteSeerAA} & \textbf{Pubmed}~\cite{Sen2008Pubmed} & \textbf{OGB-Arxiv}~\cite{hu2020ogb}  & \textbf{OGB-MAG}~\cite{hu2020ogb} \\ 
    \midrule[1pt]
    \textbf{Number of Nodes} & 2,708 & 3,327 & 19,717 & 169,343 & 1,939,743\\ 
    \textbf{Number of Edges} & 10,556 & 9,104 & 88,648 & 1,166,243 & 21,111,007\\
    \textbf{Features per Node} & 1,433 & 3,703 & 500 & 128 & 128\\
    \textbf{Number of Classes} & 7 & 6 & 3 & 23 & 349\\
    \bottomrule[2pt]
    \end{tabular}
    }
    \label{tab:dataset_statistics}
\end{table}

\section{Experimental Setup}
This section summarizes the hardware, datasets, and trajectory visualization process, including PCA-based projection~\cite{Li2017visualizing} and training of our learnable projection method.
Experiments\footnote{Code is available at \url{https://github.com/SamirMoustafa/torch-loss-landscape}} were conducted on an Intel Xeon Gold 6130 (64 cores, 256GB RAM, x86\_64 architecture). For evaluation and visualization, we used standard benchmark datasets, summarized in Table~\ref{tab:dataset_statistics}.

\paragraph{Projection Methods Pipelines and Memory Requirements:}
During training, GNN parameters from each epoch are stored to construct the matrix $D$ and train the learnable projection (Section~\ref{sec:learnable_projection}). Models are trained for up to 1000 epochs with early stopping, using ADAM (learning rate 0.01, weight decay 0.002). To ensure fair runtime comparison with PCA, our method is evaluated on CPU only, despite being GPU capable. PCA is computed via LOBPCG on CPU, and its memory cost grows with model size due to the covariance matrix.

Both methods require the matrix $D \in \mathbb{R}^{n \times d}$, but PCA additionally computes a $d \times d$ covariance matrix, while the learnable projection supports batching with batch size $B$, operating on $B \times d$ subsets of $D$. Thus, PCA incurs $\frac{d}{B}$ times the amount of memory compared to the learnable projection.

\section{Evaluation and Analysis}
Reconstruction error is computed between ground-truth GNN parameters and their reconstructions from the 2D landscape $(x, y)$. It serves as a proxy for how well the learned directions capture training dynamics. Given the difficulty of quantitatively evaluating visualization quality, reconstruction error offers an intuitive assessment metric.

\begin{figure}[t]
    \includegraphics[width=\textwidth]{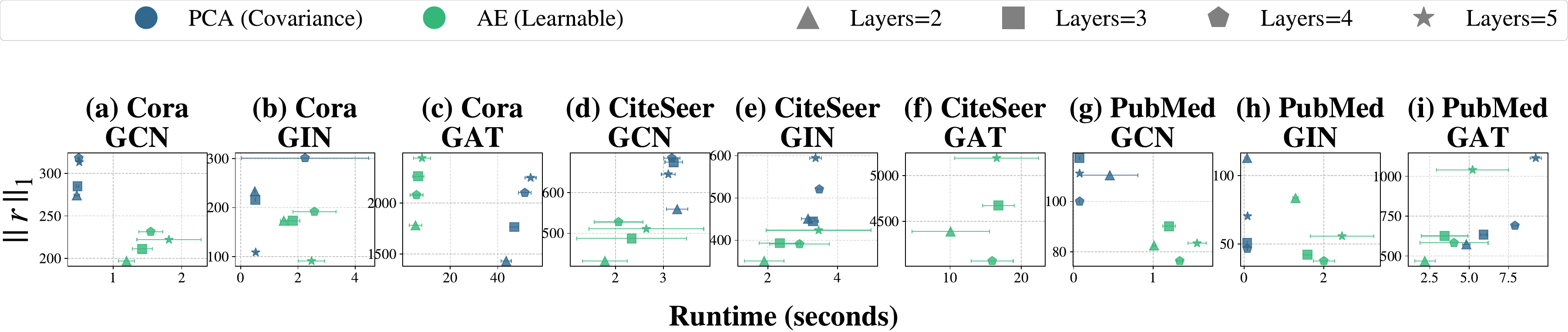}
    \caption{Reconstruction errors $\lVert r \rVert_1$ versus the time taken to compute the projection directions for GNN architectures, respectively, with $2$, $3$, $4$, and $5$ layers on different datasets. Points represent the mean of 10 runs, and the line is the standard deviation.}
    \label{fig:reconstruction_errors}
\end{figure}

Figure~\ref{fig:reconstruction_errors} shows mean reconstruction error (y-axis) and runtime (x-axis) for PCA-based projection~\cite{Li2017visualizing} and our learnable projection, averaged over ten runs.
For the GCN model, averaging over the number of layers, the learnable projection consistently lowers reconstruction error but alters runtime: on Cora the mean error drops from $297.67$ to $215.46$, on CiteSeer the error drops from $640.89$ to $489.6$, and on PubMed the error drops from $109.68$ to $83.12$, with GIN showing analogous trends. In contrast, for GAT on Cora, PCA yields slightly better reconstruction error when averaging over the number of layers ($1887.09$ vs. $2141.08$), while the learnable projection reduces runtime from $48.94$ seconds to $6.45$ seconds (7.6$\times$ faster). PCA results for GAT on CiteSeer were omitted (exceeded $256$ GB memory) due to the dense graph and GAT’s large parameter count.

\paragraph{Impact of Skip Connections and JK on Loss Landscapes:}
Figure~\ref{fig:3d_skip_connections_vs_jumping_knowledge} presents 3D loss landscape visualizations to illustrate the impact of architectural changes. For ResNet56 on CIFAR-10~\cite{KaimingDeepResidual}, the landscape without skip connections (Figure~\ref{fig:resnet_no_skip}) exhibits multiple local minima and saddle points. In contrast, adding skip connections (Figure~\ref{fig:resnet_with_skip}) smooths the landscape, mitigating vanishing gradients and facilitating optimization.
Similarly, Jumping Knowledge (JK), analogous to skip connections in GNNs, alters the loss landscape. As shown in Figure~\ref{fig:gcn_with_jk}, JK appears to reduce the non-convexity of the GCN loss surface, potentially leading to a smoother optimization landscape. Without JK (Figure~\ref{fig:gcn_no_jk}), the surface flattens and is noisy due to over-smoothing~\cite{LingfeiGNN}, obscuring the minimum.

\begin{figure}[!t]
    \footnotesize
    \centering
    \begin{subfigure}[t]{0.3\textwidth}
    \centering
    \includegraphics[width=\textwidth]{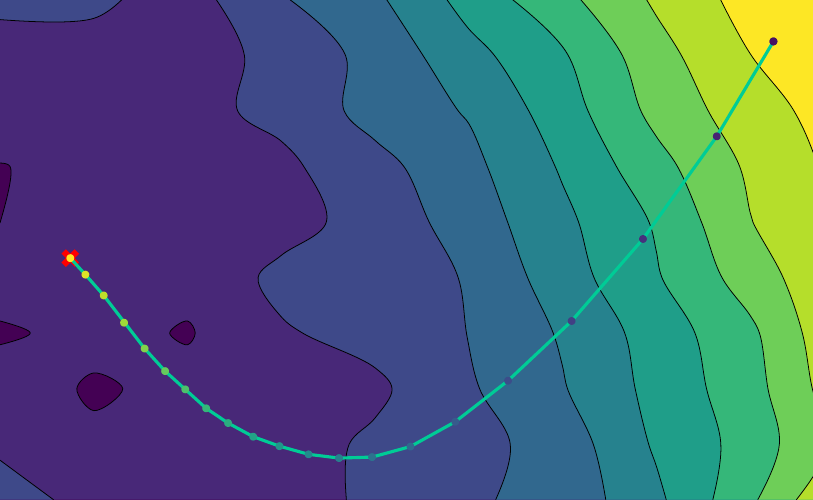}
    \caption{Five-layer GCN training trajectory on the Cora dataset shows the optimizer skipping several local minima.}
    \label{fig:trajectory_oversmoothing}
    \end{subfigure}
    \hfill
    \begin{subfigure}[t]{0.3\textwidth}
    \centering
    \includegraphics[width=\textwidth]{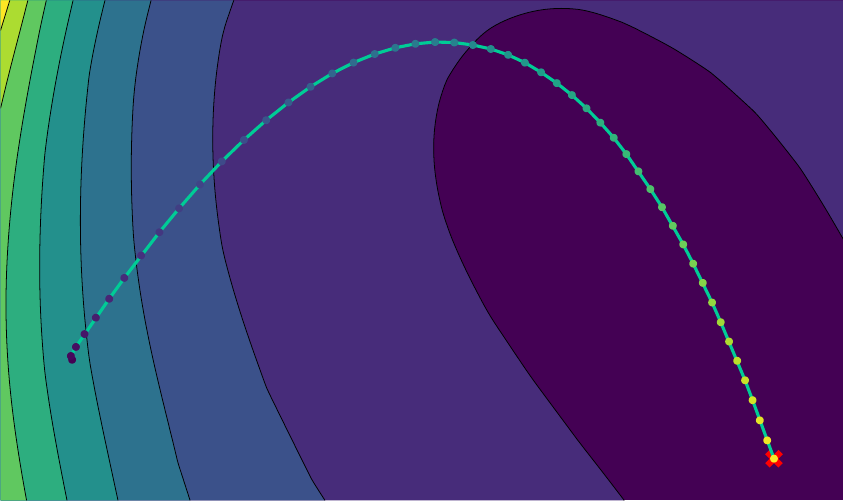}
    \caption{Training trajectory for an eight-layer GCN with residuals \cite{huang2021combining} on the OGB-Arxiv dataset.
    }
    \label{fig:trajectory_gcn_res}
    \end{subfigure}
    \hfill
    \begin{subfigure}[t]{0.3\textwidth}
    \centering
    \includegraphics[width=\textwidth]{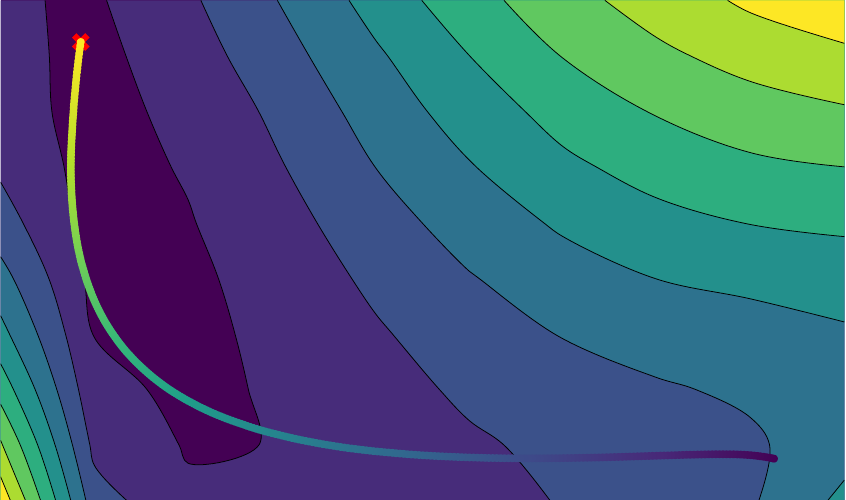}
    \caption{Trajectory for an 8-layer GAMLP during training on the OGB-MAG dataset.}
    \label{fig:trajectory_gcn_ogbn_mag}
    \end{subfigure}
    \caption{Learning trajectories of GNNs shown via the learnable projection method.}
\end{figure}

Figure~\ref{fig:trajectory_oversmoothing} illustrates the optimizer trajectory toward $\Theta^\ast$ (marked \textcolor{red}{\textbf{×}}), the final optimizer point intended to represent the optimum. However, near this identified final point, at least three local minima exist. This suboptimality arises from over-smoothing: many nodes aggregate similar information during propagation, causing their features to become indistinguishable. 

Consequently, the final GNN layer produces identical predictions for these nodes. As a result, different nodes incur the same loss value, creating a flat loss surface near the non-optimal point. This flat region disrupts optimization by having identical gradients, preventing further progress toward the true optimum.

Table~\ref{tab:cora_gcn_results} shows that increasing GCN depth degrades accuracy and increases loss, with a sharp decline beyond four layers. This supports the observation that deeper architectures induce flatter, non-convex loss landscapes. In the 6-layer case, the optimizer fails to converge to a local minimum, hindering regularization and reducing accuracy.
Jumping Knowledge variants can mitigate this degradation. \cite{huang2021combining} introduces residual connections with an error correlation mechanism that propagates residuals from training to test data. As shown in Figure~\ref{fig:trajectory_gcn_res}, their 8-layer GCN on OGB-Arxiv achieves 73.5\% test accuracy with 0.727 loss, converging to a minimum without flat surrounding regions.

\begin{wraptable}[12]{r}{0.35\textwidth} 
    \centering
    \vspace{-1.75em}
    \caption{Mean and standard deviation of loss and accuracy over 10 runs for GCN on Cora.}
    \resizebox{0.35\textwidth}{!}{
    \begin{tabular}{ccc}
    \toprule[2pt]
    \textbf{Layers} & \textbf{Loss} $\downarrow$ & \textbf{Acc. (\%)} $\uparrow$ \\
    \midrule[1pt]
    2 & 0.86 $_{\pm 0.2}$ & 81.02 $_{\pm 0.9}$ \\
    3 & \textbf{0.85} $_{\pm 0.2}$ & \textbf{81.5} $_{\pm 1.0}$ \\
    4 & 0.86 $_{\pm 0.2}$ & 80.1 $_{\pm 0.9}$ \\
    5 & 0.88 $_{\pm 0.2}$ & 79.1 $_{\pm 1.4}$ \\
    6 & 1.33 $_{\pm 0.5}$ & 77.9 $_{\pm 2.0}$ \\
    \bottomrule[2pt]
    \end{tabular}
    }
    \label{tab:cora_gcn_results}
\end{wraptable}

Figure~\ref{fig:trajectory_gcn_ogbn_mag} visualizes the loss trajectory of an 8-layer Graph Attention Multilayer Perceptron (GAMLP)~\cite{zhang2022graph} trained on OGB-MAG. Despite the model and dataset size, the learnable projection effectively visualizes the training process.

\begin{figure}[t]
    \begin{subfigure}[t]{0.3\textwidth}
        \centering
        \includegraphics[width=\textwidth]{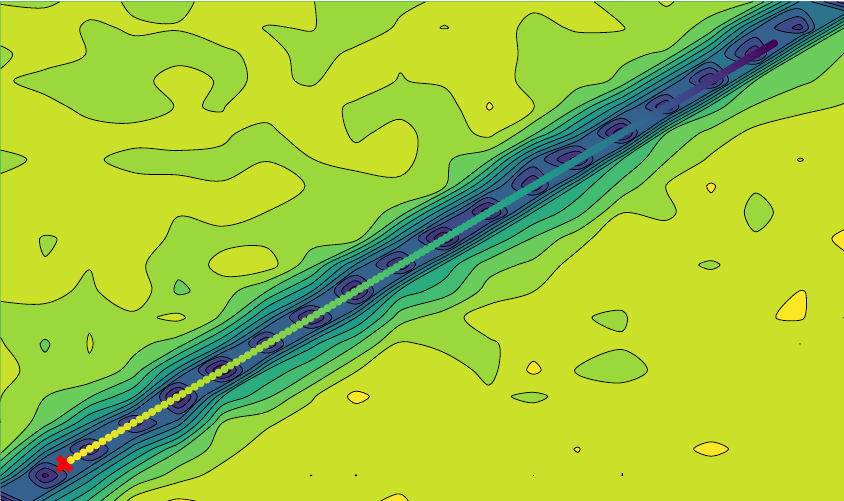}
        \caption{Training on the Cora dataset, yielding a loss of  $1.206$ and an accuracy rate of $83.6\%$.}
        \label{fig:cora_quantized_gat}
    \end{subfigure}
    \hspace{0.01\linewidth}
    \begin{subfigure}[t]{0.3\textwidth}
        \centering
        \includegraphics[width=\textwidth]{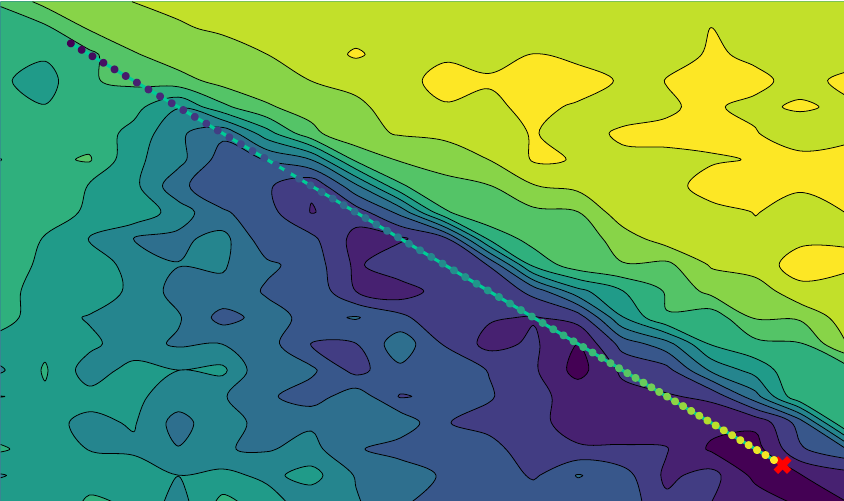}
        \caption{Training on the CiteSeer dataset, yielding a loss of  $1.09$ and \; an accuracy rate of $71.8\%$.}
        \label{fig:citeseer_quantized_gat}
    \end{subfigure}
    \hspace{0.01\linewidth}
    \begin{subfigure}[t]{0.3\textwidth}
        \centering
        \includegraphics[width=\textwidth]{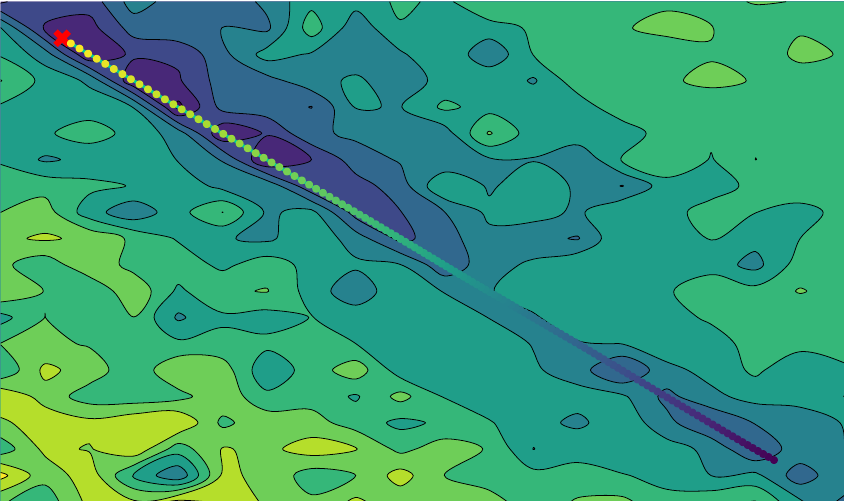}
        \caption{Training on the PubMed dataset, yielding a loss of $0.79$ and \; an accuracy rate of $75.6\%$.}
        \label{fig:pubmed_quantized_gat}
    \end{subfigure}
    \caption{Visualizing the training trajectory of a quantized GAT architecture across three different datasets, using our learnable projection method. The trajectory path appears as a straight line due to the bending in the loss landscape and its progression from a significantly far non-optimal point to the optimal point.}
    \label{fig:quantized_gat}
\end{figure}

\paragraph{Quantization and Sparsification:}
While prior work has analyzed the loss landscapes of quantized DNNs across various architectures, no study has, to our knowledge, examined this for quantized GNNs. Addressing this gap, we adopt the state-of-the-art mixed-precision GNN quantization method from~\cite{Zhu2023rmAA}, which maintains performance across diverse datasets and tasks at both node and graph levels.
Figure~\ref{fig:quantized_gat} visualizes the loss landscape of a quantized GAT, chosen for its sensitivity to perturbations in attention parameters. Quantization introduces approximation noise, increasing landscape ruggedness and local minima. Despite this, the optimizer achieves high accuracy, attributed to the method in~\cite{Zhu2023rmAA}, which uses group-wise learning rates within ADAM. However, this approach requires extensive hyperparameter tuning. Figures~\ref{fig:cora_quantized_gat} and~\ref{fig:pubmed_quantized_gat} show that, despite bypassing local minima, the model converges to accuracy-maximizing solutions.

\begin{figure}[t]
    \centering
    \begin{subfigure}[t]{0.24\textwidth}
        \includegraphics[width=\textwidth]{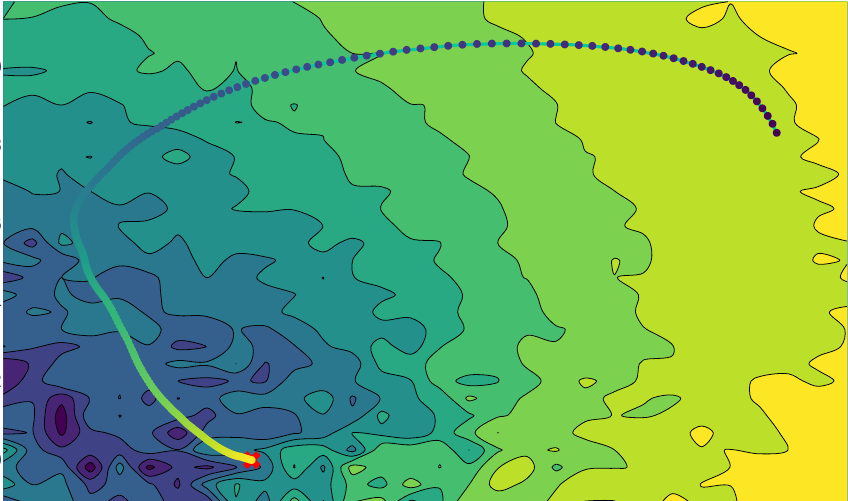}
        \subcaption{Native and complete graph structure is utilized.}
    \end{subfigure}
    \hfill
    \begin{subfigure}[t]{0.24\textwidth}
        \includegraphics[width=\textwidth]{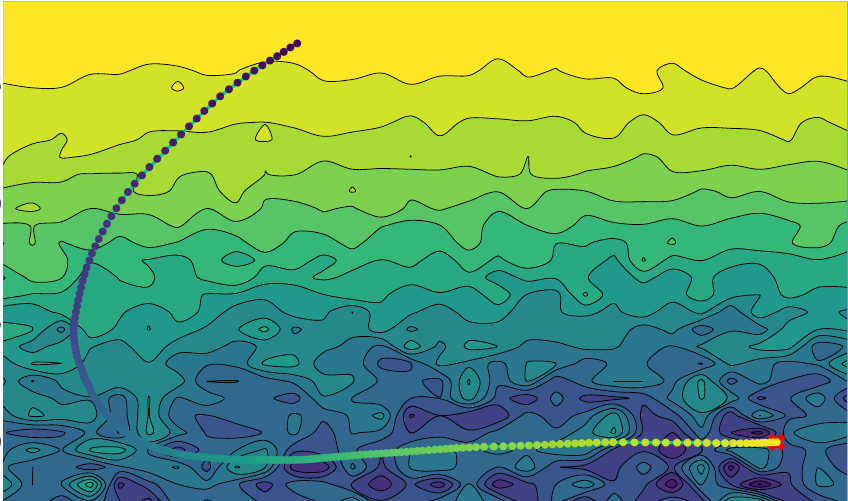}
        \subcaption{Sparsification randomly omits parts of the graph.}
        \label{fig:cora_sparsification_gin}
    \end{subfigure}
    \hfill
    \begin{subfigure}[t]{0.24\textwidth}
        \includegraphics[width=\textwidth]{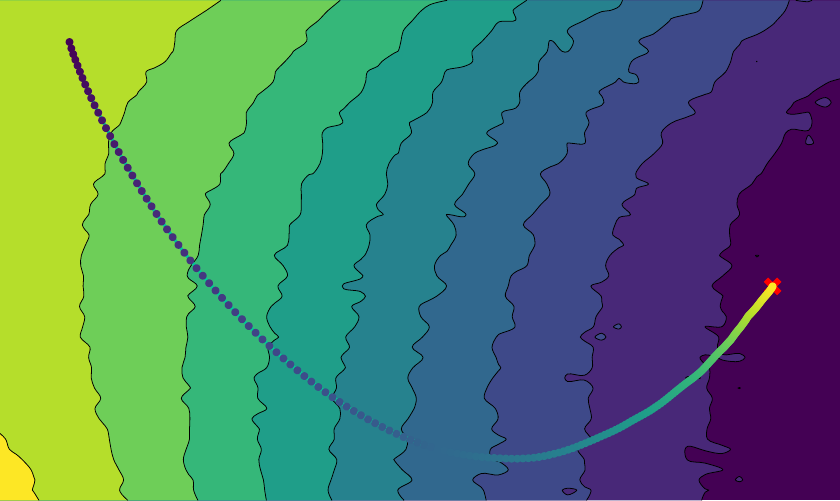}
        \subcaption{Without preconditioner, and the final loss value is $0.31$.
        }
        \label{fig:cora_gcn}
    \end{subfigure}
    \hfill
    \begin{subfigure}[t]{0.24\textwidth}
        \includegraphics[width=\textwidth]{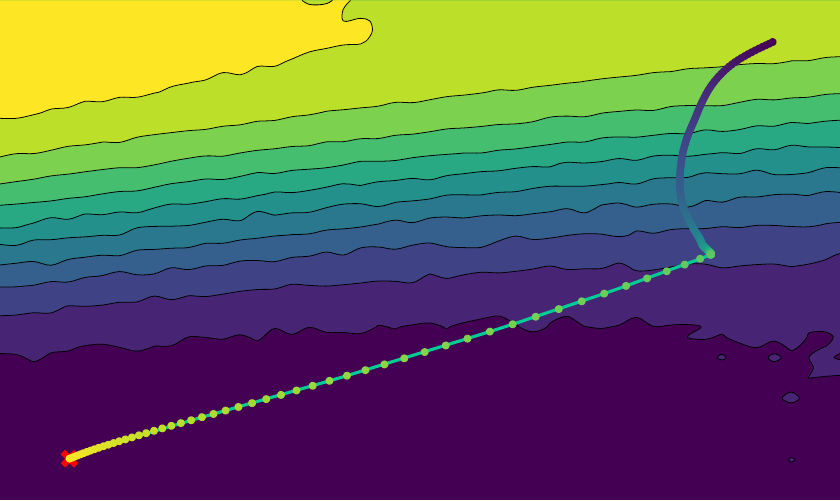}
        \subcaption{With preconditioner after the $50$-th epoch, loss is $0.03$. 
        }
        \label{fig:cora_kfac_gcn_applied}
    \end{subfigure}
    \caption{Training trajectories for GIN (a, b) and GCN (c, d) architecture during the training on Cora dataset, illustrating the effects of sparsification and preconditioning techniques.}
\end{figure}

Concerning sparsification, as noted in Section~\ref{sec:optimization}, local minima do not inherently hinder DNN training. In GNNs, graph sparsification can act as regularization, improving generalization and training efficiency by simplifying the graph and encouraging convergence to diverse local minima. We use the GIN architecture, known for matching the expressivity of the Weisfeiler-Lehman test~\cite{Xu2018a}, as its performance is highly sensitive to structural changes like sparsification, which can notably increase loss landscape ruggedness.
Figure~\ref{fig:cora_sparsification_gin} shows the impact of sparsification on the GIN loss landscape for the Cora dataset. Using GraphSAINT~\cite{graphsaint-iclr20}, we compare sparsified and non-sparsified training. Sparsification increases the number of local minima, resulting in a more rugged loss landscape.

\paragraph{KFAC Preconditioning:}
As noted in Section~\ref{sec:optimization}, the KFAC preconditioner enhances optimization by adapting step sizes to local curvature, increasing the likelihood of reaching better minima in fewer iterations. This can lower final loss and improve post-training performance.
Figure~\ref{fig:cora_kfac_gcn_applied} illustrates the impact of applying the KFAC preconditioner during the final 50 training epochs, compared to early application. This highlights its influence on the optimizer's trajectory. Notably, the pre-KFAC trajectory resembles that in Figure~\ref{fig:cora_gcn}, indicating consistent behavior prior to preconditioning.

\section{Conclusion}
We introduced a learnable projection method for visualizing GNN optimization trajectories, which outperformed the PCA-based approach in reconstruction error across most experiments. The negligible runtime increase is offset by lower memory usage, which is proportional to the batch size, enabling our method to scale to larger architectures compared to the PCA-based approach.

We analyzed the effects of over-smoothing, jumping knowledge, quantization, sparsification, and preconditioning on GNN optimization. Each impacts the loss landscape, trajectories, or both. While aligning with DNN findings, our results provide GNN-specific insights, highlighting the role of architectural choices in efficient training. Specifically, we identified that increasing GNN depth exacerbates over-smoothing, yielding flatter, non-convex loss landscapes and reduced performance. Quantization and sparsification introduce ruggedness and additional local minima. Preconditioning effectively alters optimizer trajectories, significantly aiding loss reduction.

In summary, GNN architecture, quantization, sparsification, and preconditioning substantially affect optimization trajectories, underscoring the importance of their careful consideration in GNN design and training.
\\\\
\textbf{Acknowledgment:}
Nils Kriege was supported by the Vienna Science and Technology Fund (WWTF) [10.47379/VRG19009].

\bibliographystyle{splncs04}
\bibliography{library}

\end{document}